\documentclass{article}

\usepackage[final, nonatbib]{neurips_2024}

\usepackage[utf8]{inputenc} 
\usepackage[T1]{fontenc}    
\usepackage{hyperref}       
\usepackage{url}            
\usepackage{booktabs}       
\usepackage{amsfonts}       
\usepackage{nicefrac}       
\usepackage{microtype}      
\usepackage{xcolor}  
\usepackage{float}
 \usepackage{booktabs}

\usepackage{graphicx}
\usepackage{soul,color}
\usepackage{xcolor}
\usepackage{amsmath}
\usepackage{enumitem}
\usepackage{comment}
\usepackage{amsthm}

\newcommand{\method}{xMIL-LRP }
\newcommand{\methodnospace}{xMIL-LRP}

\theoremstyle{definition}
\newtheorem{definition}{Definition}[section]

\title{xMIL: Insightful Explanations for Multiple Instance Learning in Histopathology}

\author{%
    \textbf{Julius Hense}$^{1,2,*,\dagger}$ \quad 
  \textbf{Mina Jamshidi Idaji}$^{1,2,*,\dagger}$ \quad
  \textbf{Oliver Eberle}$^{1,2}$ \quad
  \textbf{Thomas Schnake}$^{1,2}$ \\
  \textbf{Jonas Dippel}$^{1,2,3}$ \quad
  \textbf{Laure Ciernik}$^{1,2}$ \quad
  \textbf{Oliver Buchstab}$^{4}$ \quad
  \textbf{Andreas Mock}$^{4,5}$ \\
  \textbf{Frederick Klauschen}$^{1,4,5,6}$ \quad
  \textbf{Klaus-Robert Müller}$^{1,2,7,8,\dagger}$  \vspace{2mm} \\
  $^{1}$Berlin Institute for the Foundations of Learning and Data, Berlin, Germany \\
  $^{2}$Machine Learning Group, Technische Universität Berlin, Berlin, Germany \\
  $^{3}$Aignostics GmbH, Berlin, Germany \\
  $^{4}$Institute of Pathology, Ludwig Maximilian University, Munich, Germany \\
  $^{5}$German Cancer Research Center, Heidelberg, and German Cancer Consortium, Munich, Germany \\
  $^{6}$Institute of Pathology, Charité Universitätsmedizin, Berlin, Germany  \\
  $^{7}$Department of Artificial Intelligence, Korea University, Seoul, Korea \\
  $^{8}$Max-Planck Institute for Informatics, Saarbrücken, Germany \\
  $^*$Equal contribution \\
  $^{\dagger}$\texttt{\{ j.hense, mina.jamshidi.idaji, klaus-robert.mueller \}@tu-berlin.de}
}

\begin{document}

\maketitle

\begin{abstract}
Multiple instance learning (MIL) is an effective and widely used approach for weakly supervised machine learning. In histopathology, MIL models have achieved remarkable success in tasks like tumor detection, biomarker prediction, and outcome prognostication.
However, MIL explanation methods are still lagging behind, as they are limited to small bag sizes or disregard instance interactions.
We revisit MIL through the lens of explainable AI (XAI) and introduce xMIL, a refined framework with more general assumptions. We demonstrate how to obtain improved MIL explanations using layer-wise relevance propagation (LRP) and conduct extensive evaluation experiments on three toy settings and four real-world histopathology datasets.
Our approach consistently outperforms previous explanation attempts with particularly improved faithfulness scores on challenging biomarker prediction tasks. Finally, we showcase how xMIL explanations enable pathologists to extract insights from MIL models, representing a significant advance for knowledge discovery and model debugging in digital histopathology. Codes are available at: \href{https://github.com/bifold-pathomics/xMIL}{https://github.com/bifold-pathomics/xMIL}.
\end{abstract}
\raggedbottom

\section{Introduction}  \label{sec:intro}


Multiple instance learning (MIL) \cite{DIETTERICH199731, NIPS1997_82965d4e} is a learning paradigm in which a single label is predicted from a bag of instances. Various MIL methods have been proposed, differing in how they aggregate instances into bag information \cite{ilse2018attentionmil, shao2021transmil, li2021dualstreammil, lu2021clam, sharma2021clustermil, yang2022remixmil, zhang2022dtfdmil, fillioux2023s4mil, fourkioti24camil, bilal2023aggregation}. MIL has become particularly popular in histopathology, where gigapixel microscopy slides are cut into patches representing small tissue regions. From these patches, MIL models can learn to detect tumor \cite{campanella2019clinical} or classify disease subtypes \cite{lu2021clam}, aiming to support pathologists in their routine diagnostic workflows. They have further demonstrated remarkable success at tasks that even pathologists cannot perform reliably due to a lack of known histopathological patterns associated with the target, e.g., predicting clinically relevant biomarkers \cite{kather2020pan, echle2021biomarkers, arslan2024systematic} or outcomes like survival \cite{saillard2020survival, skrede2020outcome} directly from whole slide images.


Explaining which visual features a MIL model uses for its prediction is highly relevant in this context. It allows experts to sanity-check the model strategy \cite{lapuschkin2019unmasking}, e.g., whether a model focuses on the disease area for making a diagnosis. This is particularly important in histopathology, where models operating in high-stake environments are prone to learning confounding factors like artifacts or staining differences instead of actual signal \cite{howard2021signatures, hagele2020resolving, histo-xai-review}. On top of that, MIL explanations can enable pathologists to discover novel connections between visual features and prediction targets. For example, the explanations could reveal a previously unknown association of a histopathological pattern with poor survival, leading to the identification of a targetable disease mechanism. Previous works have shown the potential of scientific knowledge discovery from explainable AI (XAI) \cite{histo-xai-review, Binder2021MorphologicalAM, schweizer2023analysing, schutt2017quantum, keith2021combining}.

\begin{figure}[t]
    \centering
    \includegraphics[width=0.8\textwidth]{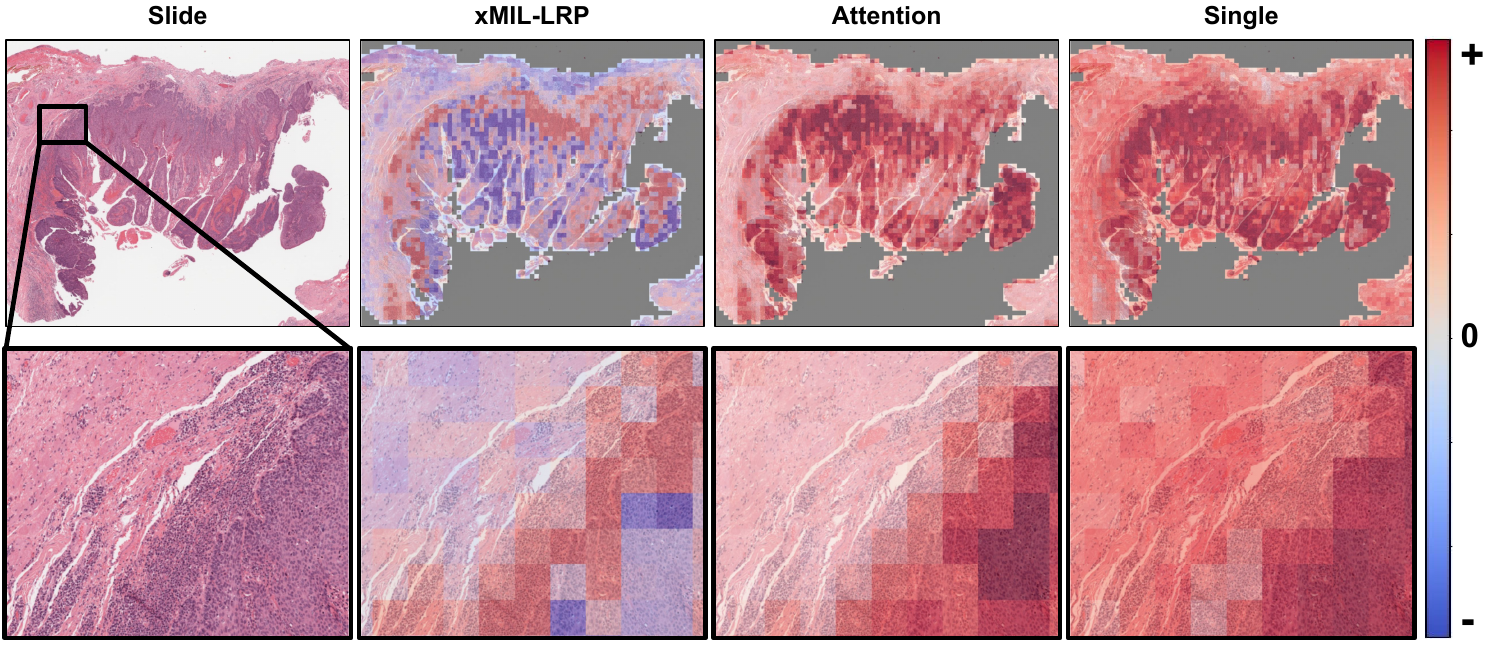}
    \caption{In digital pathology, heatmaps guide the identification of tissue slide areas most important for a model prediction. The figure displays heatmaps from different MIL explanation methods (columns) for a head and neck tumor slide (top row) with a selected zoomed-in region (bottom row). The MIL model has been trained to predict HPV status. The \method heatmap shows that the model identified evidence in favor of an HPV infection at the tumor border (red area) and evidence against an HPV infection inside the tumor (blue area, lower half of the tissue). The dominant blue region explains why the model mispredicted the slide as HPV-negative. Investigation of the tumor border by a pathologist revealed a higher lymphocyte density, which is one of the known recurrent but not always defining visual features of HPV infection in head and neck tumors. \method allows pathologists to extract fine-grained insights about the model strategy. In contrast, the ``attention'' and ``single'' methods neither explain the negative prediction nor distinguish the relevant areas.}
    \label{fig:hnscc_heatmap_main}
\end{figure}

Most studies have used attention scores as MIL explanations \cite{ilse2018attentionmil, shao2021transmil, lu2021clam, lipkova2022deep, lu2021ai, wagner2023transformer}. However, it has been shown that attention heatmaps are limited in faithfully reflecting model predictions \cite{wiegreffe-pinter-2019-attention, jain-wallace-2019-attention, ali2022xai, javed2022additivemil}. Further MIL explanation methods have been proposed, including perturbation schemes passing modified bags through the model \cite{early2022shapmil} and architectural changes towards fully additive MIL models \cite{javed2022additivemil}. Nevertheless, these methods do not account for the complexities inherent to many histopathological prediction tasks, as they are limited to small bag sizes or disregard instance interactions.

We revisit MIL through the lens of XAI and introduce xMIL, a more general and realistic multiple instance learning framework including requirements for good explanations. We then present \methodnospace, an adaptation of layer-wise relevance propagation (LRP) \cite{bach-plos15, montavon2019layer} to MIL. \method distinguishes between positive and negative evidence, disentangles instance interactions, and scales to large bag sizes. It applies to various MIL models without requiring architecture modifications, including Attention MIL \cite{ilse2018attentionmil} and TransMIL \cite{shao2021transmil}. We assess the performance of multiple explanation techniques via three toy experiments, which can serve as a novel benchmarking tool for MIL explanations in complex tasks with instance interactions and context-sensitive targets. We further perform faithfulness experiments on four real-world histopathology datasets covering tumor detection, disease subtyping, and biomarker prediction. \method consistently outperforms previous attempts across all tasks and model architectures, with the biggest advantages observed for Transformer-based biomarker prediction.

Figure \ref{fig:hnscc_heatmap_main} showcases the importance of understanding positive and negative evidence for a prediction.
Only \method uncovers that the model found evidence for the presence of the biomarker, but stronger evidence against it. This explains why it predicted the biomarker to be absent and enables pathologists to extract insights about the visual features that support or reject the presence of the biomarker according to the model. 
The example illustrates the strength of our approach, suggesting that \method represents a significant advance for model debugging and knowledge discovery in histopathology.

The paper is structured as follows: In Section \ref{sec:background}, we review MIL assumptions, models, and explanation methods related to this work. In Section \ref{sec:towards_xMIL}, we introduce xMIL as a general form of MIL, and \method as a solution for it. In Section \ref{sec:experiments}, we experimentally show the improved explanation quality of our approach. We demonstrate how to extract insights from example heatmaps in Section
\ref{sec:usecases}. Our contributions are summarized as follows:
\begin{itemize}[leftmargin=9pt]
    \item \textbf{Methodical}: Despite attempts to apply XAI to MIL models in histopathology (e.g.\ \cite{lu2021clam, lipkova2022deep, lu2021ai,  wagner2023transformer, javed2022additivemil, early2022shapmil, chen2022pan, calderaro2023deep, song2024analysis, el2024regression}), there exists no formalism guiding the interpretation of the heatmaps and defining their desired properties. xMIL is a novel framework addressing this gap. Within xMIL, heatmaps estimate the instances’ impact on the bag label, which makes their interpretation straightforward and insightful.

    \item  \textbf{Empirical}: Our extensive empirical evaluation of XAI methods for MIL on synthetic and real-world histopathology datasets is the first of its kind. It reveals that the widely used MIL explanation methods regularly yield misleading results. In contrast, \method sets a new state-of-the-art for explainability in AttnMIL and TransMIL models in histopathology.

    \item \textbf{Insight generation}: Previous studies \cite{javed2022additivemil, early2022shapmil} conducted qualitative assessments of heatmaps on easy-to-learn datasets like CAMELYON or TCGA NSCLC. The insights gained in these settings are limited to model debugging, i.e., ``Does the model focus on the disease area?'' To our knowledge, we are the first to present a method generating heatmaps that enable pathologists to extract fine-grained insights about the model in a difficult biomarker prediction task.
\end{itemize}

\section{Background} \label{sec:background}

\subsection{Multiple instance learning (MIL)} \label{sec:MIL}

\textbf{MIL formulations}. In MIL, a sample is represented by a bag of instances $X = \{ \mathbf{x}_1, \cdots, \mathbf{x}_K \}$ with a bag label $y$, where $\mathbf{x}_k \in \mathbb{R}^{D}$ is the $k$-th instance. The number of instances per bag $K$ may vary across samples. In its standard formulation \cite{DIETTERICH199731, NIPS1997_82965d4e, ilse2018attentionmil}, the instances of a bag exhibit neither dependency nor ordering among each other. It is further assumed that binary instance labels $y_{k} \in \{ 0, 1 \}$ exist but are not necessarily known. The binary bag label is $1$ if and only if at least one instance label is $1$, i.e., $y = \max_{k} \{ {y_{k}} \}$. Various extensions have been proposed \cite{carbonneau2018milsurvey, foulds2010milassumptions}, each making different assumptions about the relationships between instances and bag labels.

\textbf{MIL models}. MIL architectures typically consist of three components as illustrated in Figure~\ref{fig:xMIL_blockdiagram}: a backbone extracting instance representations, an aggregation function fusing the instance representations into a bag representation, and a prediction head inferring the final bag prediction. As recent foundation models for histopathology have become powerful feature extractors suitable for a wide range of tasks \cite{wagner2023transformer, wang2022ctranspath, dippel2024rudolfv,vorontsov2023virchow,chen2024uni}, the weights of the backbone are often frozen, allowing for a more efficient training. For aggregation, earlier works used parameter-free mean or max pooling approaches \cite{Pinheiro2015, Fengpooling2017, Wentaopooling2017}. Recently, attention mechanisms could improve performance, flexibly extracting relevant instance-level information using non-linear weighting \cite{ilse2018attentionmil,lu2021clam, pappas2017explicit} and self-attention \cite{shao2021transmil, kernelmil2021}. Attention MIL (AttnMIL) \cite{ilse2018attentionmil} computes a weighted average of the instances' feature vectors via a single attention head. TransMIL \cite{shao2021transmil} uses a custom two-layer Transformer architecture, viewing instance representations as tokens. The bag representation is extracted from the class token at the final layer. TransMIL allows for computing arbitrary pairwise interactions between all instances relevant to the prediction task. While various extensions of AttnMIL and TransMIL have been proposed (e.g., \cite{li2021dualstreammil, lu2021clam, sharma2021clustermil, yang2022remixmil, zhang2022dtfdmil, rymarczyk2022protomil, struski2023promil, fillioux2023s4mil, fourkioti24camil, bilal2023aggregation}), these two methods are arguably prototypical and among the most commonly used in the digital histopathology community.

\textbf{MIL explanation methods}. From the few studies investigating MIL interpretability, most of them use attention heatmaps \cite{ilse2018attentionmil, shao2021transmil, lu2021clam, lipkova2022deep, lu2021ai, wagner2023transformer}.
Moreover, basic gradient- and propagation-based methods have been explored for specific architectures and applications \cite{pirovano2020gradmil,sadafi2023pixelmil}. Sadafi et al.\ \cite{sadafi2023pixelmil} applied LRP to generate pixel-level attributions for single-cell images in a blood cancer diagnosis task, but did not consider its potential for instance-level explanations. Perturbation-based methods, building on model-agnostic approaches like SHAP \cite{lundberg2017shap}, perturb bag instances and compute importance scores from the resulting change in the model prediction; Early et al.\ \cite{early2022shapmil} proposed passing bags of single instances through the model (``single''), dropping single instances from bags (``one-removed''), and sampling coalitions of instances to be removed (``MILLI''). Javed et al.\ \cite{javed2022additivemil} introduced ``additive MIL'', providing directly interpretable instance scores while constraining the model's ability to capture instance interactions.

\subsection{Limitations of MIL in histopathology} \label{sec:mil_limitations}

Histopathological datasets and prediction tasks are diverse and come with various inherent challenges. We highlight the following three features.
\begin{itemize}[leftmargin=9pt]
    \item
    \textbf{Instance ambiguity}. Instances are small high-resolution patches from large images. Their individual information content may be limited, as they can be subject to noise or only be interpretable as part of a larger structure. For example, it is not always possible to distinguish a benign high-grade adenoma from a malignant adenocarcinoma on a patch level due to their similar morphology.
    \item 
    \textbf{Positive, negative, and class-wise evidence}. A single bag may contain evidence for multiple classes that a MIL model needs to weigh for correct decision-making. In survival prediction, for example, a strong immune response may support longer survival, while an aggressive tumor pattern speaks for shorter survival.
    \item 
    \textbf{Instance interactions}. In many prediction tasks, it may be necessary to consider interactions between instances. A gene mutation may generate morphological alterations in the tumor area, the tumor microenvironment, and the healthy tissue, all of which may need to be considered together to reliably predict the biomarker.
\end{itemize}
Existing MIL formulations make explicit assumptions about the relationship between instances and bag labels \cite{foulds2010milassumptions}, limiting their ability to capture the full complexity of a histopathological prediction task.
The standard MIL formulation, in particular, does not consider any of the aforementioned aspects, rendering it an unsuitable framework for most histopathological settings.

Similarly, previous MIL explanation methods suffer from various shortcomings that limit their applicability in real-world histopathology datasets. The direct interpretability of attention scores is insufficient to faithfully reflect the model predictions \cite{wiegreffe-pinter-2019-attention, jain-wallace-2019-attention, ali2022xai}. Moreover, they cannot distinguish between positive, negative, or class-wise evidence \cite{javed2022additivemil}. Purely gradient-based explanations may suffer from shattered gradients, resulting in unreliable explanations \cite{DBLP:conf/icml/BalduzziFLLMM17}. Perturbation-based approaches come with high computational complexity. While the linear ``single'' and ``one removed'' methods require $K$ forward passes per bag, MILLI scales quadratically with the number of instances \cite{early2022shapmil}. In histopathology, where bags typically contain more than 1,000 and frequently more than 10,000 instances, quadratic runtime is practically infeasible. 
Additive MIL and linear perturbation-based methods do not consider higher-order instance interactions. In prediction tasks depending on interactions, linear perturbation-based explanations may fail to provide faithful explanations, while additive models may not achieve competitive performances.

\section{Methods} \label{sec:towards_xMIL}
\begin{figure}[t!]
    \centering
    \includegraphics[width=0.8\textwidth]{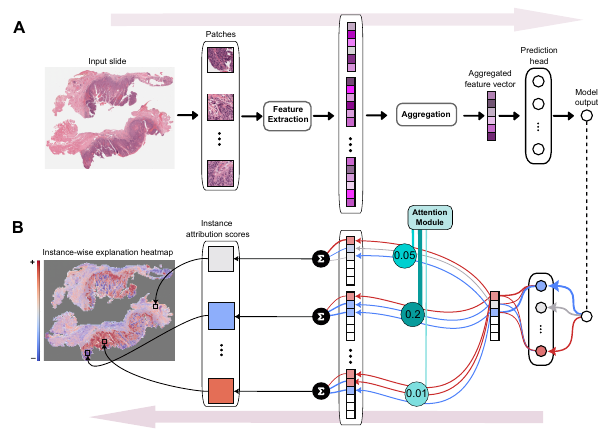}
    \caption{The two steps of xMIL: estimating the aggregation function (A) and the evidence function (B). Panel A shows a block diagram of a MIL model applied to a histopathology slide. The feature extraction module is typically a combination of a frozen foundation model followed by a  shallow MLP. In most of the recent MIL models, the aggregation module uses attention mechanisms for combining the instance feature vectors into a single feature representation per bag. The prediction head is a linear layer or an MLP. Panel B schematically shows \method for explaining AttnMIL. In \methodnospace, the model output is backpropagated to the input instances. The colored lines represent the relevance flow. Red and blue colors encode the positive and negative values. The attention module is handled via the AH-rule as described in Section \ref{sec:xmil-lrp}. As discussed in Section \ref{sec:xmil_properties}, the instance explanation scores can be computed at the output of the foundation model or at the input level.}
    \label{fig:xMIL_blockdiagram}
\end{figure}

\textbf{Notation}. We denote vectors with boldface lowercase letters (e.g., $\mathbf{x}$), scalars with lowercase letters (e.g., $x$), and sets with uppercase letters (e.g., $X$).

\subsection{xMIL: An XAI-based framework for multiple instance learning} \label{sec:xMIL}

We address the limitations discussed in Section \ref{sec:mil_limitations} and introduce a more general formulation of MIL: explainable multiple instance learning (\textbf{xMIL}). At its core, we propose moving away from the notion of instance labels towards context-aware \textit{evidence scores}, which better reflect the intricacies of histopathology while laying the foundation for developing and evaluating MIL explanation methods.

\begin{definition}[Explainable multiple instance learning]
Let $X=\{ \mathbf{x}_1, \dots, \mathbf{x}_K \}$ be a bag of instances with a bag label $y \in \mathbb{R}$. \\
(1) There exists an \textit{aggregation function} $\mathcal{A}$ that maps the bag to its label, i.e., $\mathcal{A}(X) = y$. We make no assumptions about the relationship among the instances or between the instances and the label $y$. \\
(2) There exists an \textit{evidence function} $\mathcal{E}$ assigning an \textit{evidence score} $\mathcal{E}(X, y, \mathbf{x}_k) = \epsilon_k \in \mathbb{R}$ to any instance $\mathbf{x}_{k}$ in the bag, quantifying the impact the instance has on the bag label $y$. \\
The aim of xMIL is to estimate (i) the aggregation function $\mathcal{A}$ and (ii) the evidence function $\mathcal{E}$.
\end{definition}

\begin{definition}[Properties of the evidence function] \label{def:evidence_function}
Let $\mathbf{x}_{k}, \mathbf{x}_{k'}$ be instances from a bag $X$. We assume that $\mathcal{E}$ has the following properties. \\
(1) \textit{Context sensitivity}. The evidence score $\epsilon_{k}$ of instance $\mathbf{x}_{k}$ may depend on other instances from $X$. \\
(2) \textit{Positive and negative evidence}. If $\epsilon_{k} > 0$, the instance $\mathbf{x}_{k}$ has a positive impact on the bag label $y$. If $\epsilon_{k} < 0$, then $\mathbf{x}_{k}$ has a negative impact on $y$. If $\epsilon_{k} = 0$, then $\mathbf{x}_{k}$ is irrelevant to $y$. \\
(3) \textit{Ordering}. If $\epsilon_{k} > \epsilon_{k'} \ge 0$, then instance $\mathbf{x}_{k}$ has a higher positive impact on $y$ than $\mathbf{x}_{k'}$. If $0 \ge \epsilon_{k'} > \epsilon_{k}$, then instance $\mathbf{x}_{k}$ has a higher negative impact on $y$ than $\mathbf{x}_{k'}$.
\end{definition}

Similar to our definition, previous works described context sensitivity and accounting for positive and negative evidence as desirable properties of MIL explanation methods \cite{javed2022additivemil, early2022shapmil}. However, xMIL integrates these principles directly into the formalization of the MIL problem. 


In contrast to previous MIL formulations, xMIL addresses the potential complexities within histopathological prediction tasks by refraining from posing strict assumptions on $\mathcal{A}$. Via the evidence function $\mathcal{E}$, we suggest that instances may vary in their ability to support or refute a class and that their influence may depend on the context within the bag. In practice, the evidence function is often unknown, as the notion of an ``impact'' on the bag label is hard to quantify. For the standard MIL setting, however, the binary instance labels fulfill the criteria of the evidence function. Therefore, xMIL is a more general and realistic formulation of multiple instance learning for histopathology.

We can learn the aggregation function $\mathcal{A}$ via training a MIL model. To gain deeper insights into the prediction task by estimating the evidence function $\mathcal{E}$, we design an explanation method for the learned aggregation function with characteristics suitable to the properties of the evidence function.

\subsection{xMIL-LRP: Estimating the evidence function} \label{sec:xmil-lrp}

We introduce \method as an efficient solution to xMIL, bringing layer-wise relevance propagation (LRP) to MIL.
LRP is a well-established XAI method \cite{bach-plos15, montavon2018methods} with a large body of literature supporting its performance in explaining various types of architectures in different tasks \cite{ali2022xai, montavon2019layer, Arras2019, eberle2020building, schnake2021higher, letzgus2022}. Starting from the prediction score of a selected class, the LRP attribution of neuron $i$ in layer $l$ receives incoming messages from neurons $j$ from subsequent layer $l+1$, resulting in relevance scores $r_i^{(l)} = \sum_j \frac{q_{ij}}{\sum_{i'} q_{i'j}}\cdot r_j^{(l+1)}$, 
with $q_{ij}$ being the contribution of neuron $i$ of layer $l$ to relevance $r_j^{(l+1)}$ \footnote{In some XAI research papers, uppercase letters, e.g.,  $R_j^{(l)}$, are used for denoting relevance values.}. A variety of so-called ``propagation rules'' have been proposed \cite{montavon2019layer} to specify the contribution $q_{ij}$ in specific model layers. For the attention mechanism, as a core component of many MIL architectures, we employ the AH-rule introduced by Ali et al.\ \cite{ali2022xai}. In a general attention mechanism, let $\mathbf{z}_k=\left[z_{kd}\right]_d$ be the embedding vector of the $k$-th token and $p_{kj}$ the attention score between tokens $k$ and $j$. The output vector of the attention module is $\mathbf{y}_j = \sum_{k} p_{kj} \mathbf{z}_k$. The AH-rule of LRP treats attention scores as a constant weighting matrix during the backpropagation pass of LRP. If $R(y_{jd})$ is the relevance of the $d$-th dimension of $\mathbf{y}_j=\left[y_{jd}\right]_d$, the AH-rule computes the relevance of the $d$-th feature of $\mathbf{z}_k$ as: 
\begin{equation}
    R(z_{kd})=\sum_{j} \frac{z_{kd} p_{kj}}{\sum_{i} z_{id}p_{ij}} R(y_{jd}).
\end{equation}
This formulation can be directly applied to AttnMIL, and also adapted to a QKV attention block in a transformer, where $\mathbf{z}_k$ is the embedding associated with the value representation.

We illustrate the effect of this rule in AttnMIL in Figure \ref{fig:xMIL_blockdiagram}-B. The relevance flow separates the instances weighted by the attention mechanism into positive, negative, and neutral instances, resulting in more descriptive heatmaps that better show the relevant tissue regions compared to attention scores.

We further implement the LRP-$\epsilon$ rule for linear layers followed by ReLU activation function \cite{montavon2018methods}, as well as the LN-rule to address the break of conservation in layer norm \cite{ali2022xai}, with details presented in Appendix \ref{app:lrp_details}.

At the instance-level, \method assigns each instance $\mathbf{x}_k=[x_{kd}]_d \in \mathbb{R}^D$ a relevance vector $\mathbf{r}_k=[r_{kd}]_d$ with $r_{kd}=R(x_{kd})=r^{(0)}_{kd}$ being the relevance score of the $d$-th feature of $\mathbf{x}_k$. We define the instance-wise relevance score as an estimate for the evidence score of the instance as $\hat{\epsilon}_k = \sum_d r_{kd}$.

\subsection{Properties of \method and other explanation methods} \label{sec:xmil_properties}

The properties of \method are particularly suitable for estimating the evidence function:


\textbf{Context sensitivity}: \method disentangles instance interactions and contextual information as it jointly considers the relevance flow across the whole bag. LRP and Gradient $\times$ Input (G$\times$I) are rooted in a deep Taylor decomposition of the model prediction \cite{montavon-pr17} and consequently 
capture dependencies between features by tracing relevance flow through the components of the MIL model. While attention is context-aware, it is limited to considering dependencies of features at a specific layer. The ``single'' method is unaware of context. ``One-removed'' and additive MIL can only capture the impact of individual instances on the prediction.


\textbf{Positive and negative evidence}: \method relevance scores are real-valued and can identify whether an instance supports or refutes the model prediction. Features irrelevant to the prediction will receive an explanation score close to zero.
Therefore, the range of explanation scores matches the range of the assumed evidence function. The same holds for additive MIL, MILLI, and ``one-removed''. 
Attention and ``single'' do not distinguish between positive and negative evidence.

\textbf{Conservation}: Following the conservation principle of LRP, \method provides an instance-wise decomposition of the model output, i.e., $\sum_k \hat{\epsilon}_k=\sum_{k,d} r_{kd} = y$. This instance-level conservation also holds for additive MIL, but not for the other discussed methods. The local conservation principle of LRP \cite{montavon2019layer} further allows us to analyze attribution scores at the instance feature vector level without requiring propagation through the foundation model---the instance-wise attribution scores are the same at any layer of the model.






\section{Experiments and results} \label{sec:experiments}

\textbf{Baseline methods}. We compared several explanation methods to our \method (see Appendix \ref{app:baseline_xAI} for details). For AttnMIL and TransMIL, we selected Gradient$\,\times\,$Input (\textbf{G$\times$I}) \cite{baehrens10a,DBLP:journals/corr/ShrikumarGSK16} and Integrated gradients (\textbf{IG}) \cite{sundararajan2017axiomatic} as gradient-based baselines. We further included the ``single'' perturbation method (\textbf{single}) \cite{early2022shapmil}, which involves using predictions for individual instances as explanation scores. Single is the only computationally feasible perturbation-based approach for the bag sizes considered here (up to 24,000).
We evaluated raw attention scores for AttnMIL and attention rollout \cite{abnar-zuidema-2020-quantifying} for TransMIL (\textbf{attn}). In the random baseline (\textbf{rand}), instance scores were randomly sampled from a standard normal distribution. For additive attention MIL (AddMIL) \cite{javed2022additivemil}, we assessed raw attention scores (\textbf{attn}) and the model-intrinsic instance-wise predictions (\textbf{logits}).

\subsection{Toy experiments} \label{sec:toy_experiments}


We designed novel toy experiments to assess and compare the characteristics of \method and the baseline methods for AttnMIL, TransMIL, and AddMIL in controlled settings. We focused on evaluating to what extent the explanations account for \textit{context sensitivity} and \textit{positive and negative evidence}, i.e., the first two characteristics of the evidence function according to Definition \ref{def:evidence_function}, which we consider crucial aspects for explaining real-world histopathology prediction tasks.

Inspired by previous works \cite{ilse2018attentionmil, early2022shapmil}, we sampled bags of MNIST images \cite{deng2012mnist}, with each instance representing a number between \textbf{0} and \textbf{9}. We defined three MIL tasks for these bags:
\begin{itemize}[leftmargin=9pt]
    \item 
    \textbf{4-Bags}: The bag label is class 1 if \textbf{8} is in the bag, class 2 if \textbf{9} is in the bag, class 3 if \textbf{8} and \textbf{9} are in the bag, and class 0 otherwise. The dataset was proposed by Early et al.\ \cite{early2022shapmil}. In this setting, the model needs to learn basic instance interactions.
    \item 
    \textbf{Pos-Neg}: We define \textbf{4}, \textbf{6}, \textbf{8} as positive and \textbf{5}, \textbf{7}, \textbf{9} as negative numbers. The bag label is class 1 if the amount of unique positive numbers is strictly greater than that of unique negative numbers, and class 0 otherwise. The model needs to adequately weigh positive and negative evidence to make correct predictions.
   \item 
   \textbf{Adjacent Pairs}: The bag label is class 1 if it contains any pair of consecutive numbers between \textbf{0} and \textbf{4}, i.e., (\textbf{0},\textbf{1}), (\textbf{1},\textbf{2}), (\textbf{2},\textbf{3}) or (\textbf{3},\textbf{4}), and class 0 otherwise. In this case, the impact of an instance is contextual, as it depends on the presence or absence of adjacent numbers.
\end{itemize}

To assess the explanation quality, we first defined valid \textit{evidence scores} as ground truths according to Definition \ref{def:evidence_function}. For each dataset, we require one evidence function per predicted class $c$, denoted by $\mathcal{E}^{(c)}(X, y, \mathbf{x}_k) = \epsilon_{k}^{(c)}$. We assigned $\epsilon_{k}^{(c)} = 1$ if $\mathbf{x}_k$ supports class $c$, $\epsilon_{k}^{(c)} = -1$ if the instance refutes class $c$, and $\epsilon_{k}^{(c)} = 0$ if it is irrelevant. We aimed to measure whether an explanation method correctly distinguishes instances with positive, neutral, and negative evidence scores. Therefore, we computed a two-class averaged area under the precision-recall curve (AUPRC-2), measuring if the positive instances received the highest and the negative instances the lowest explanation scores. We assessed AttnMIL and TransMIL models and repeated each experiment 30 times. The details of the ground truth, the evaluation metric, and the experimental setup are provided in Appendix \ref{app:toy_experiments_details}.

Table \ref{tab:toy_experiments} displays the test AUROC scores of the three models across datasets, demonstrating that the models solve the tasks to varying degrees, alongside the performances of the explanation methods. We find that \method outperformed the other explanation approaches across MIL models and datasets in all but one setting. It reached particularly high AUPRC-2 scores in the 4-Bags and Pos-Neg datasets while being most robust in the more difficult Adjacent Pairs setting. Attention severely suffered from the presence of positive and negative evidence, which it cannot distinguish by design. While IG performed comparably to \method for AttnMIL models, it was inferior for TransMIL. Notably, the test AUROC of AddMIL was worse in all settings, resulting in explanations that are not competitive with the post-hoc explanation methods on AttnMIL and TransMIL. This supports our point that AddMIL may not perform competitively in difficult prediction tasks. The single perturbation method provided good explanations in the Pos-Neg setting, where numbers have a fixed evidence score irrespective of the other instances in the bag. However, in 4-Bags and Adjacent Pairs, the method's performance decreased, as it always assigns the same score to the same instance regardless of the bag context. In contrast, \method is both context-sensitive and identifies positive and negative instances. Since we expect that these aspects are common features of many real-world histopathological datasets, we conclude that our method is the only suitable approach for such complex settings.

 \setlength{\tabcolsep}{3.75pt}
\begin{table}[t!]
\scriptsize
\caption{Results of the toy experiments. We report AUPRC-2 scores of MIL explanation methods on three toy datasets measuring how well a method identified instances with positive and negative evidence scores (mean ± std.\ over 30 repetitions). The highest mean scores are bold and the second highest are underlined. We also display the model performances ("Test AUROC", mean ± std.).}
\label{tab:toy_experiments}
\centering
\begin{tabular}{lccccccccc}
\toprule
{} & \multicolumn{3}{c}{4-Bags} & \multicolumn{3}{c}{Pos-Neg} & \multicolumn{3}{c}{Adjacent Pairs} \\
\midrule
{} & AttnMIL & TransMIL & AddMIL & AttnMIL & TransMIL & AddMIL & AttnMIL & TransMIL & AddMIL \\
Test AUROC   &  1.00 ± 0.00 &  1.00 ± 0.00 & 0.98 ± 0.02     & 0.97 ± 0.00 &   0.98 ± 0.00 &  0.89 ± 0.03	  &   0.88 ± 0.08 &  0.92 ± 0.06 & 0.77 ± 0.07 \\
\midrule
Rand    &   0.31 ± 0.00 & 0.31 ± 0.00 &  -- & 0.42 ± 0.00 &  0.42 ± 0.00 &   -- &  0.54 ± 0.00 &  0.54 ± 0.00 & -- \\
Attn &    0.53 ± 0.00 &   0.52 ± 0.00  &  0.54 ± 0.01  &    0.45 ± 0.00 &  0.46 ± 0.03 & 0.48 ± 0.02	&  0.61 ± 0.01 &  0.60 ± 0.01 & 0.63 ± 0.04  \\
Single    &  0.87 ± 0.02 &     \underline{0.85 ± 0.07} & --    & 0.89 ± 0.00 &   \underline{0.91 ± 0.02} & --    &    0.73 ± 0.06 & \underline{0.77 ± 0.06} & -- \\
Logits    & -- &         -- & 0.79 ± 0.11       & -- &      -- & 0.68 ± 0.17       &    -- &            -- & 0.71 ± 0.09 \\
G$\times$I       &       0.72 ± 0.08 &               0.40 ± 0.07  & -- &     0.72 ± 0.14 &            0.44 ± 0.06 & -- &            0.63 ± 0.05 &      0.57 ± 0.05 & -- \\
IG        &       \underline{0.88 ± 0.01} &               0.80 ± 0.09  & -- &     \textbf{0.93 ± 0.00} &            0.82 ± 0.08 & -- &            \underline{0.75 ± 0.03} &      0.72 ± 0.07 & -- \\
xMIL-LRP   &         \textbf{0.91 ± 0.01} &           \textbf{0.90 ± 0.02} & --  &      \underline{0.91 ± 0.01} &       \textbf{0.92 ± 0.01} & -- &            \textbf{0.77 ± 0.04} &             \textbf{0.81 ± 0.04} & -- \\
\bottomrule
\end{tabular}
\end{table}

\subsection{Histopathology experiments}
\label{sec:histo_experiments}

\textbf{Datasets and model training.} To evaluate the performance of explanations on real-world histopathology prediction tasks, we considered four diverse datasets of increasing task difficulty covering tumor detection, disease subtyping, and biomarker prediction. These datasets had previously been used for benchmarking in multiple studies \cite{bilal2023aggregation, javed2022additivemil, chen2024uni, wang2021heal}.
\begin{itemize}[leftmargin=9pt]
    \item 
    CAMELYON16 \cite{bejnordi2017camelyon} consists of 400 sentinel lymph node slides, of which 160 carry to-be-recognized metastatic lesions of different sizes. It is a well-established tumor detection dataset.
    \item 
    The TCGA NSCLC dataset (abbreviated as NSCLC) contains 529 slides with lung adenocarcinoma (LUAD) and 512 with lung squamous cell carcinoma (LUSC). The prediction task is to distinguish these two non-small cell lung cancer (NSCLC) subtypes.
    \item 
    The TCGA HNSC HPV dataset \cite{bilal2023aggregation} (abbreviated as HNSC HPV) has 433 slides of head and neck squamous cell carcinoma (HNSC). 43 of them were affected by a human papillomavirus (HPV) infection diagnosed via additional testing \cite{campbell2018tcgascc}. HPV infection is an essential biomarker guiding prognosis and treatment \cite{bilal2023aggregation}. The task is to identify the HPV status directly from the slides. Label imbalances and the complexity of the predictive signature are key challenges in this task.
    \item 
    The TCGA LUAD TP53 (abbreviated as LUAD TP53) dataset contains 529 lung adenocarcinoma (LUAD) slides, 263 of which exhibit a mutation of the TP53 gene, which is one of the most common mutations across cancers. In lung cancer, it is associated with poorer prognosis and resistance to chemotherapy and radiation \cite{mogi2011tp53}. Previous works showed that TP53 mutation can be predicted from LUAD slides \cite{wang2021heal, coudray2018nsclc}.
\end{itemize}


We generated patches at 20x magnification and obtained 10,454 $\pm$ 6,236 patches per slide across all datasets (mean $\pm$ std.).
Features were extracted using the pre-trained CTransPath \cite{wang2022ctranspath} foundation model and aggregated using AttnMIL or TransMIL.\footnote{We did not include AddMIL in the real-world experiments, as it is difficult to compare heatmaps from different models without having a ground truth like in the toy experiments (Section \ref{sec:toy_experiments}). Also notice that faithfulness evaluations are not applicable, since AddMIL explanations are faithful by design \cite{javed2022additivemil}.}
Additional details regarding the datasets and training procedure are described in Appendix \ref{app:histo_details}.

We report the mean and standard deviation of the test set AUROC over 5 repetitions in Table~\ref{tab:dropping_results}.
In all but one case, TransMIL outperformed AttnMIL, with the largest margin observed in the difficult TP53 dataset. Our results generally align with performances reported in previous works \cite{bilal2023aggregation, chen2024uni, wang2021heal}.

 \setlength{\tabcolsep}{3.75pt}

\begin{table}[t!]
\scriptsize
\caption{Results of the faithfulness experiments. AUPC values per dataset, MIL model, and explanation method (mean ± std.\ over all slides). \emph{Lower scores indicate higher faithfulness}. The best performance per setting (significant minimum based on the paired t-tests) is highlighted in bold. We also display the model performances (``Test AUROC'', mean ± std.\ over 5 repetitions).}
\label{tab:dropping_results}
\centering
\begin{tabular}{lcccc|cccc}
\toprule
{} & \multicolumn{4}{c}{AttnMIL} & \multicolumn{4}{c}{TransMIL}  \\
\toprule
{} & CAMELYON16 & NSCLC & HNSC HPV & LUAD TP53 & CAMELYON16 & NSCLC & HNSC HPV & LUAD TP53 \\
Test AUROC & 0.93 ± 0.00 & 0.95 ± 0.00 & 0.88 ± 0.06 & 0.71 ± 0.01 & 0.95 ± 0.01 & 0.96 ± 0.00 & 0.88 ± 0.05 & 0.75 ± 0.01 \\
\midrule

Rand & 0.94 ± 0.13 & 0.98 ± 0.04 & 0.97 ± 0.07 & 0.84 ± 0.14 & 0.95 ± 0.11 & 0.98 ± 0.08 & 1.00 ± 0.01 & 0.94 ± 0.17 \\

Attn & 0.65 ± 0.46 & 0.70 ± 0.27 & 0.94 ± 0.18 & 0.65 ± 0.14 & 0.63 ± 0.45 & 0.91 ± 0.22 & 0.95 ± 0.15 & 0.64 ± 0.38 \\

Single & 0.61 ± 0.43 & 0.42 ± 0.26 & 0.78 ± 0.23 & 0.34 ± 0.16 & 0.42 ± 0.35 & 0.53 ± 0.26 & 0.92 ± 0.13 & 0.73 ± 0.33 \\

G$\times$I & 0.92 ± 0.19 & 0.81 ± 0.35 & 0.81 ± 0.25 & 0.44 ± 0.23 & 0.82 ± 0.36 & 0.79 ± 0.30 & 0.87 ± 0.20 & 0.66 ± 0.40 \\

IG & 0.62 ± 0.44 & 0.75 ± 0.38 & 0.78 ± 0.25 & 0.38 ± 0.20 & 0.88 ± 0.23 & 0.99 ± 0.01 & 1.00 ± 0.00 & 0.99 ± 0.01 \\

\method & \textbf{0.51 ± 0.38} & \textbf{0.25 ± 0.22} & \textbf{0.71 ± 0.24} & \textbf{0.31 ± 0.16} & \textbf{0.29 ± 0.30} & \textbf{0.45 ± 0.26} & \textbf{0.75 ± 0.23} & \textbf{0.24 ± 0.28} \\

\bottomrule
\end{tabular}
\end{table}

\textbf{Faithfulness evaluation.} As the evidence functions $\mathcal{E}$ of our histopathology datasets are unknown, we resorted to assessing \textit{faithfulness}, i.e., how accurately explanation scores reflect the model prediction \cite{samek2016evaluating, blucher2024flipping}. The primary goal of the faithfulness experiments is to evaluate the ordering of relevance scores (Property 3 of the evidence function in Definition \ref{def:evidence_function}). Faithfulness can be quantified by progressively excluding instances from the most relevant first (MORF) to the least relevant last and measuring the change in prediction score. The area under the resulting perturbation curve (AUPC) indicates how faithfully the identified ordering of the instances affects the model prediction. The lower the AUPC score, the more faithful the method. We calculated AUPC for correctly classified slides. Further methodological details are provided in Appendix \ref{app:perturbation_details}. 

In Figure \ref{fig:transmil_dropping_results}, we show the perturbation curves and AUPC boxplots for the patch-dropping experiment for TransMIL in our four datasets (Figure \ref{fig:attnmil_dropping_results} shows the results for AttnMIL). Additionally, we summarize our results in Table \ref{tab:dropping_results}. To test the difference in the AUPC values among the baseline explanation methods, we performed paired t-tests between the random baseline vs.\ all methods and \method vs. all other baselines. The p-values were corrected using the Bonferroni method for multiple comparison correction. All tests resulted in significant differences except for random baseline vs.\ G$\times$I for CAMELYON16 and attention for HNSC HPV.

\method significantly achieved the lowest average AUPC compared to the baselines, providing the most faithful explanations across all tasks and model architectures. Especially evident with the TransMIL model, \method accurately decomposed the mixing of patch information via self-attention. Notably, the largest margin of \method to other methods could be observed in the more challenging biomarker prediction tasks of the HNSC HPV and LUAD TP53 datasets.

The results also reflect whether the explanation scores contain meaningful positive/negative evidence for the target class (Property 2 of the evidence function in Definition \ref{def:evidence_function}): if so, we expect the model’s prediction to flip when all patches supporting the target class are excluded. In Figure \ref{fig:transmil_dropping_results}, the model decision always flips when patches are excluded based on \method scores, whereas other methods show inconsistent results.

Attention scores, as the most widely used explanation approach for MIL in histopathology, did not provide faithful explanations outside the simple tumor detection setting in the CAMELYON16 dataset. This remarkably highlights their limited usefulness as model explanations and confirms previously reported results in other domains \cite{wiegreffe-pinter-2019-attention, jain-wallace-2019-attention, ali2022xai}.
Passing single instances through the model (``single'') achieved good faithfulness scores for simpler tasks and AttnMIL, but performed worse for Transformer-based biomarker prediction.

\begin{figure}[t!]
    \centering
    \includegraphics[width=0.95\textwidth]{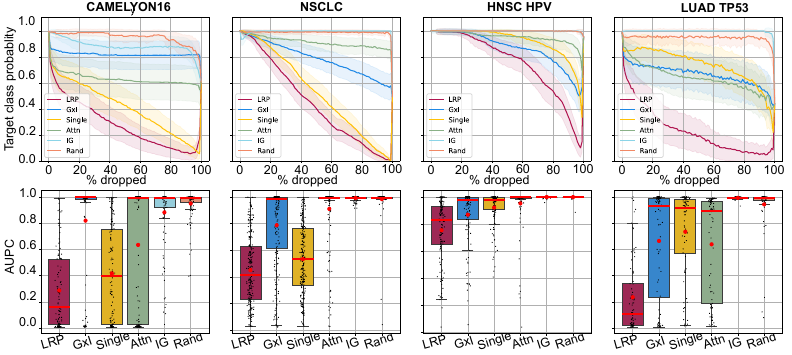}
    \caption{Patch dropping results for TransMIL. The first row depicts the perturbation curves, where the solid lines are the average perturbation curve and the shaded area is the standard error of the mean at each perturbation step. Each boxplot on the second row shows the distribution of AUPC values for all test set slides per explanation methods. In each boxplot, the red line marks the median and the red dot marks the mean. \emph{Lower perturbation curves and AUPCs represent higher faithfulness.}}
    \label{fig:transmil_dropping_results}
\end{figure}

\section{Extracting insights from \method heatmaps} \label{sec:usecases}

The identification of predictive features for HPV infection in head and neck carcinoma from histopathological slides is a challenging task for pathologists. In this task, there are partially known morphological patterns associated with the class label. We provide a brief overview of the known histological features differentiating HPV-negative and HPV-positive HNSC in Appendix \ref{app:additional_heatmaps} and Figure \ref{fig:hnscc_hpv_features}.  In the following, we demonstrate how faithful \method explanations can support pathologists in gaining insights about the model strategy and inferring task-relevant features.

We extracted explanation scores for the best-performing TransMIL models. To increase the readability of resulting heatmaps, we clipped the scores per slide at the whiskers of their boxplots, which extended 1.5 times the interquartile range from the first and third quartiles. We then translated them into a zero-centered red-blue color map, with red indicating positive and blue negative scores. Notice that the explanation methods operate on different scales. For \methodnospace, a positive relevance score indicates support for the explained label, while a negative score contradicts it.

We revisit the example of the HNSC tumor with a false-positive prediction of an HPV infection in Figure \ref{fig:hnscc_heatmap_main}. As previously noted, only \method indicates that the model recognizes evidence of HPV infection in the tumor border, but not the remaining tumor.
Despite a prediction score close to 0, all relevance scores from the single method were between 0.95--0.97, suggesting that context-free single-instance bags may not be informative in this task. We observed this phenomenon across various slides.

Heatmaps of additional examples are provided in Appendix \ref{app:additional_heatmaps}. In Figure \ref{fig:hnscc_heatmap_3}, \method accurately delineates and distinguishes HPV-positive tumor islands from the surrounding stroma. In this simple case, attention also provides a reasonable explanation. Figure \ref{fig:hnscc_heatmap_1} presents another correctly classified HPV-positive sample. Here, \method outlines spatially consistent slide regions with clear positive evidence, distinct from regions of negative or mixed evidence (top row). Most notably, the subepithelial mucous glands (bottom row), which are not associated with HPV, are correctly highlighted in blue, unlike in the attention map. In Figure \ref{fig:hnscc_heatmap_4}, we display a false positive slide. In this case, \method allowed us to identify that the evidence of HPV-positivity can be attributed to an unusual morphology of an HPV-negative tumor that shares some morphological features usually associated with HPV infection (e.g., smaller tumor cells with hyperchromatic nuclei, dense lymphocyte infiltrates). 

\section{Conclusion}
We introduced xMIL, a more general and realistic MIL framework for histopathology, formalizing requirements for MIL explanations via the evidence function.
We adapted LRP to MIL as \methodnospace, experimentally demonstrated its advantages over previous explanation approaches, and showed how access to faithful explanations can enable pathologists to extract insights from a biomarker prediction model.
Thus, xMIL is a step toward increasing the reliability of clinical ML systems and driving medical knowledge discovery, particularly in histopathology. Despite being motivated by the challenges in histopathology, our approach presented here can be directly transferred to other problem settings that require explaining complex MIL models, e.g., in video, audio, or text domains. Furthermore, a detailed analysis of potentially complex dependencies between instances, especially in the context of multi-modal inputs, represents a promising direction for future research.

\section*{Acknowledgements}
The results shown here are in whole or part based upon data generated by the TCGA Research Network: https://www.cancer.gov/tcga. This work was in part supported by the German Ministry for Education and Research (BMBF) under Grants 01IS14013A-E, 01GQ1115, 01GQ0850, 01IS18025A, 031L0207D, 01IS18037A, and BIFOLD24B and by the Institute of Information \& Communications Technology Planning \& Evaluation (IITP) grants funded by the Korea government (MSIT) (No. 2019-0-00079, Artificial Intelligence Graduate School Program, Korea University and No. 2022-0-00984, Development of Artificial Intelligence Technology for Personalized Plug-and-Play Explanation and Verification of Explanation). 



\bibliographystyle{unsrt}
\bibliography{bibliography}


\newpage

\appendix

\section{Appendix}

\subsection{Baseline MIL explanation methods}\label{app:baseline_xAI}

\paragraph*{Attention maps}

Attention scores have commonly been used as an explanation of the model by considering attention heatmaps, assuming that they reflect the importance of input features \cite{vaswani_attention2017}. 

In AttnMIL, a bag representation is computed as an attention-weighted average of instance-level representations, i.e.,
\begin{align} \label{eq:attnmil}
    g(X) = \sum_{k=1}^{K} a_k f(\mathbf{x}_k), \quad a_{k} = \text{softmax} \left( \mathbf{w}^T \left( \tanh(V f(\mathbf{x}_k)^T) \odot \text{sigm}(U f(\mathbf{x}_k)^T) \right) \right)
\end{align}
where $f(\mathbf{x}_k)$ is an instance and $g(X)$ the bag representation. The attention scores $a_k$ assign to each patch an attribution with $0 \leq a_k \leq 1$, and have been used as instance-wise explanation scores \cite{ilse2018attentionmil}.

In TransMIL, the attention heads deliver self-attention vectors $\mathbf{A}_{h}^{l} \in \mathbb{R}^{(K + 1) \times (K + 1)}$ for each head $h$ and Transformer layer $l$, recalling that the first token is the class token. Mean pooling is often used for fusing the self-attention matrices of different heads, i.e., $\mathbf{A}^l = \left \langle \mathbf{A}_h^l \right \rangle_h$. The attention scores from the class token to the instance tokens can be used as attribution scores, i.e., $\mathbf{A}^l_{(1, 2:)}$. Alternatively, attention rollout has been proposed to summarize the self-attention matrices over layers \cite{abnar-zuidema-2020-quantifying}. For a model with $L$ Transformer layers, attention rollout combines $\{\mathbf{A}^l\}_{l=1}^{L}$ as $\tilde{\mathbf{A}}=\prod_{l=1}^{L} \check{\mathbf{A}}^l$ where $\check{\mathbf{A}}^l=0.5\mathbf{A}^l + 0.5\mathbf{I}$, with $\mathbf{I}$ being the identity matrix. Then, similar to the layer-wise attention scores, the heatmap is defined as the attention rollout of the class token to the instances, i.e.,  $\tilde{\mathbf{A}}_{(1, 2:)}$.

\paragraph*{Gradient-based methods}

Gradient-based methods utilize gradient information to derive feature relevance scores. Pirovano et el.\ \cite{pirovano2020gradmil} combined raw gradients to identify the most relevant features and derive tiles that activate these features the most.

Various other gradient-based methods have been proposed in the XAI literature, including saliency maps and Gradient$\,\times\,$Input (G$\times$I) \cite{baehrens10a,DBLP:journals/corr/ShrikumarGSK16} and Integrated Gradients (IG) \cite{sundararajan2017axiomatic}. These methods can easily be adapted to compute explanations in MIL. We obtain the gradient of a MIL model prediction $\hat{y}$ with respect to a patch $\nabla \hat{y}(\mathbf{x}_k)$. 

For G$\times$I, we can then define the relevance score of the $k$-th instance as $\sum_d [\nabla \hat{y}(\mathbf{x}_k)]_d x_{kd}$, with $x_{kd}$ being the $d$-th feature of $\mathbf{x}_k$. 

Integrated gradients (IG) \cite{sundararajan2017axiomatic} computes the gradients of the model's output with respect to the input, integrated over a path from a baseline to the actual input. The baseline is typically set to zero, and so we do. The explanation score of the $k$-th instance is computed as $\sum_d \text{IG}(x_{kd})$, where the relevance score of the $d$-th feature of the $k$-th instance $\text{IG}(x_{kd})$ is computed as  
\begin{equation}
    \text{IG}(x_{kd}) = x_{kd} \cdot \int_{\alpha=0}^{1} \frac{f( \alpha \mathbf{X})}{\partial x_{kd}} d\alpha,
\end{equation}

where $f$ is the model and $\mathbf{X}$ is the $K\times D$ feature matrix of the bag (with $K$ being the number of instances and $D$ being the number of features for each instance).  We used the implementation of IG available in Captum \cite{kokhlikyan2020captum} with the internal batch size set to the number of instances in a bag.

\paragraph*{Perturbation-based methods}

The idea of perturbation-based explanation methods is to perturb selected instances of a bag and derive importance scores from the resulting change in the model prediction. It builds on model-agnostic post-hoc local interpretability methods like LIME \cite{ribeiro2016lime} and SHAP \cite{lundberg2017shap}.

Early et al.\ \cite{early2022shapmil} proposed and evaluated multiple perturbation-based methods of different complexity. The ``single'' method passes bags of single patches $X_k = \{ \mathbf{x}_{k} \}$ for $k=1,\ldots,K$ through the model and uses the outcome $f(X_k)$ as explanation score. ``One removed'' drops single patches, i.e.\, constructs bags $\check{X}_{k} = X \backslash X_k$ for $k=1,\ldots,K$ and defines the difference to the original prediction score $f(X) - f(\check{X}_k)$ as explanation. The ``combined'' approach takes the mean of these two scores. As these methods cannot account for patch interactions, Early et al.\ also propose an algorithm to sample coalitions of patches to be perturbed, called MILLI. They show that MILLI outperforms the baselines on toy datasets when instance interactions need to be considered. The complexity of MILLI is $O(nK^2)$, where $n$ is the number of coalitions and K is the bag size.

\paragraph*{Additive MIL}

The idea of additive MIL \cite{javed2022additivemil} is to make the MIL model inherently interpretable by designing the bag-level prediction to be a sum of individual instance predictions. Let function $f$ be a feature extractor and $\psi_{m}, \psi_{p}$ MLPs. In many cases, particularly for Attention MIL \cite{ilse2018attentionmil}, a MIL model $p$ can be written as
\begin{align}
    p(X) = \psi_{p}\left( \sum_{k=1}^{K} a_{k} f\left( \mathbf{x}_{k} \right) \right) \quad \text{with} \quad a_{k}=\text{softmax}_{k}(\psi_{m}(X)),
\end{align}
where $a_k$ is the attention score of instance $k$. For Attention MIL, $\psi_{m}$ is defined as the inner part of the softmax function of Equation \ref{eq:attnmil}, and $\psi_{p}$ as prediction head outputting class logits. To obtain an additive model, the authors suggest to instead compute
\begin{align}
    p(X) = \sum_{k=1}^{K} \psi_{p}\left( a_{k} f\left( \mathbf{x}_{k} \right) \right).
\end{align}
This way, the bag prediction becomes the sum of the individual instance predictions $\psi_{p}\left( a_{k} f\left( \mathbf{x}_{k} \right) \right)$, which can be used as instance explanation scores. These instance logits are proportional to the Shapley values of the instances \cite{javed2022additivemil}. In our experiments, we consider the proposed additive variant of Attention MIL (AddMIL).

\subsection{Layer-wise Relevance Propagation (LRP)}\label{app:lrp_details}

LRP is a method for explaining neural network predictions by redistributing the output's relevance back through the network to the input features. The redistribution follows a relevance conservation principle, where the total relevance of each layer is preserved as it propagates backward. If $r_j^{(l)}$ denotes the relevance of neuron $j$ in layer $l$, conservation means that $\sum_{j} r_j^{(l_1)} = \sum_{i} r_i^{(l_2)}$ holds for any two layers $l_1$ and $l_2$. As a general principle, LRP posits
\begin{equation}
    r_i^{(l)} = \sum_j \frac{q_{ij}}{\sum_{i'} q_{i'j}}\cdot r_j^{(l+1)},
\end{equation}
with $q_{ij}$ being the contribution of neuron $i$ of layer $l$ relevance $r_j^{(l+1)}$. There are ``propagation rules'' for various layer types \cite{montavon2019layer, montavon2018methods} that specify $q_{ij}$ for different setups.

\textbf{Feed forward neural network}. The following generic rule holds for propagating relevance through linear layers followed by ReLU \cite{montavon2019layer}:
\begin{equation} \label{eq:linear_lrp_rule}
    r_i^{(l)} = \sum_j \frac{a_j \rho(w_{ij})}{\epsilon + \sum_{i'} a_{i'} \rho(w_{i'j})}\cdot r_j^{(l+1)},
\end{equation}
where $a_{j}$ is the activation of neuron $j$ in layer $l$, $w_{ij}$ the weight from neuron $i$ of layer $l$ to neuron $j$ of layer $l+1$, $\epsilon$ a stabilizing term to prevent numerical instabilities, and $\rho(w_{ij})$ a modification of the weights of the linear layer. For example, if $\rho(w_{ij})=w_{ij} + \gamma \text{max}(w_{ij}, 0)$, then Equation \ref{eq:linear_lrp_rule} is called LRP-$\gamma$ rule. For $\gamma=0$, this equation is called LRP-$\epsilon$ rule.

\textbf{LayerNorm}. Assume $\mathbf{z}_k$ is the embedding of the $k$-th token and $\mathbf{y}_k=\text{LayerNorm}(\mathbf{z}_k)$ as:
\begin{equation}
    \mathbf{y}_k=\frac{\mathbf{z}_k - \text{E}\{\mathbf{z}\}}{\text{std}\{\mathbf{z}\} + \epsilon},
\end{equation}
where $\text{E}\{\mathbf{z}\}$ and $\text{std}\{\mathbf{z}\}$ are the expected values and standard deviation of the tokens. 

For propagating relevance through LayerNorm, Ali et al.\ \cite{ali2022xai} suggested the LN-rule as the following:
\begin{equation}
     R(z_{kd}) = \sum_{j} \frac{z_{kd} (\delta_{kj} - \frac{1}{N})}{\sum_{i} z_{id} (\delta_{ij} - \frac{1}{N})} R(y_{jd}),
\end{equation}
where $\delta_{kj} = \begin{cases} 1, & \text{if } k = j \\ 0, & \text{otherwise} \end{cases}$ and  $z_{kd}$ is the $d$-th dimension of $\mathbf{z}_k$ and $R(z_{kd})$ is the relevance assigned to it. In practice, LN-rule is implemented by detaching $\text{std}\{\mathbf{z}\}$ and handling it as a constant.
\subsection{Toy experiments: Training and evaluation details} \label{app:toy_experiments_details}

\textbf{Evidence functions.} We define the evidence functions for the three datasets as follows. We write $\mathbf{x}_k \sim \textbf{n}$ to indicate that instance $k$ represents MNIST number $n$.
\begin{itemize}[leftmargin=9pt]
    \item
    In the \textit{4-Bags} dataset, \textbf{8} supports classes 1 and 3 but refutes classes 0 and 2, while \textbf{9} supports classes 2 and 3 but refutes classes 0 and 1. Hence, for $\mathbf{x}_k \sim \textbf{8}$, we define $\epsilon_k^{(c)} = 1$ for $c \in \{ 1, 3 \}$ and  $\epsilon_k^{(c)} = -1$ for $c \in \{ 0, 2 \}$. For $\mathbf{x}_k \sim \textbf{9}$, we set $\epsilon_k^{(c)} = 1$ for $c \in \{ 2, 3 \}$ and $\epsilon_k^{(c)} = -1$ for $c \in \{ 0, 1 \}$. In all other cases, $\epsilon_{k}^{(c)} = 0$.
    \item 
    In \textit{Pos-Neg}, \textbf{4}, \textbf{6}, and \textbf{8} instances support class 1 and refute class 0, and vice versa for \textbf{5}, \textbf{7}, \textbf{9}. Hence, we set $\epsilon_k^{(1)} = 1$ and $\epsilon_k^{(0)} = -1$ if $\mathbf{x}_k \sim \{ \textbf{4}, \textbf{6}, \textbf{8} \}$, $\epsilon_k^{(1)} = -1$ and $\epsilon_k^{(0)} = 1$ if $\mathbf{x}_k \sim \{ \textbf{5}, \textbf{7}, \textbf{9} \}$, and $\epsilon_{k}^{(c)} = 0$ otherwise.
    \item 
    In \textit{Adjacent Pairs}, \textbf{4} supports class 1 and refutes class 0 if \textbf{3} is also present, but is irrelevant otherwise. That is, for $\mathbf{x}_k \sim \textbf{4}$, we set $\epsilon_k^{(1)} = 1$ and $\epsilon_k^{(0)} = -1$ if \textbf{3} is also in in the bag, and $\epsilon_k^{(0)} = 0$ otherwise. The evidence scores for the other numbers are defined accordingly.
\end{itemize}

\textbf{Evaluation metric.} We aim to measure whether an explanation method correctly distinguishes instances with positive, neutral, and negative evidence scores. We separate this into two steps: quantify the separation between positive and non-positive instances, and quantify the separation between negative and non-negative instances. Let $\mathbf{e}^{(c)} = [ \epsilon^{(c)}_1, \dots, \epsilon^{(c)}_K ]$ be the evidence scores for some bag $X = \{ \mathbf{x}_1, \dots, \mathbf{x}_K \}$ and class $c$, and $\mathbf{s}^{(c)} = [ \hat{\epsilon}^{(c)}_1, \dots, \hat{\epsilon}^{(c)}_K ]$ be the class-wise explanation scores from some explanation method. We define $\mathbf{e}^{(c)}_{pos} = \min(\mathbf{e}^{(c)}, \mathbf{0})$ and $\mathbf{e}^{(c)}_{neg} = \min(-\mathbf{e}^{(c)}, \mathbf{0})$ as the binarized positive and negative evidence, and compute
\begin{align}
    \text{AUPRC-2} &= \frac{1}{2} \cdot \left( \text{AUPRC}( \mathbf{e}^{(c)}_{pos}, \mathbf{s}^{(c)}) + \text{AUPRC}(\mathbf{e}^{(c)}_{neg}, -\mathbf{s}^{(c)}) \right).
\end{align}
We utilize the area under the precision-recall curve (AUPRC) to account for potential imbalances. Our AUPRC-2 metric can be interpreted as the one-vs-all AUPRC score for detecting positive and negative instances. It becomes $1$ if all instances with positive / negative evidence have been assigned the highest / lowest evidence scores. For each dataset and explanation method, we computed the AUPRC-2 across all classes and test bags and report the average score.

\textbf{Experimental details.} Instead of training end-to-end MIL models, we obtained feature vectors with 512 dimensions for each MNIST image via a ResNet18 model pre-trained on Imagenet from the TorchVision library \cite{torchvision2016}. For each bag, we first sampled a subset of numbers, where each number was selected with a probability of 0.5, and then randomly drew 30 MNIST feature vectors from this subset. We used 2,000 bags for training, 500 for validation, and 1,000 for testing. We trained AttnMIL and TransMIL models with a learning rate of 0.0001 for a maximum of 1000 and 200 epochs for AttnMIL and TransMIL, respectively. We finally selected the model with the lowest validation loss. We repeated each model training 30 times and report means and standard deviations across repetitions. Each experiment with its repetitions was run on single CPUs in less than 24 hours, respectively. We used the same setting for training AddMIL, but with an Adam optimizer as in the original paper \cite{javed2022additivemil}.

\subsection{Histopathology experiments: Data and training details}
\label{app:histo_details}

\textbf{Dataset details.} We downloaded TCGA HNSC, LUAD, and LUSC datasets from TCGA website. The HPV status of HNSC dataset and the TP53 mutations of LUAD dataset were downloaded from cBioPortal \cite{cerami2012cbio, gao2013integrative, de2023analysis}. We applied the following splits.
\begin{itemize}[leftmargin=9pt]
    \item 
    CAMELYON16: We used the pre-defined test set of 130 slides, and randomly split the remaining slides into 230 for training and 40 for validation. 
    \item 
    NSCLC: As in previous works \cite{shao2021transmil, javed2022additivemil}, we randomly split the slides into 60\% training, 15\% validation, and 25\% test data.
    \item 
    HNSC HPV: Due to the low number of HPV-positive samples, we uniformly split the dataset into three cross-validation folds like in previous work \cite{bilal2023aggregation}.
    \item 
    LUAD TP53: We randomly split the slides into 60\% training, 15\% validation, and 25\% test data.
\end{itemize}

\textbf{Preprocessing details.} We extracted patches from the slides of $256 \times 256$ pixels without overlap at 20x magnification (0.5 microns per pixel). We identified and excluded background patches via Otsu's method \cite{otsu1975threshold} on slide thumbnails and applied a patch-level minimum standard deviation of 8.

\textbf{Training details.} For training, we sampled bags of 2048 patches per slide and passed their feature representations through the MIL model. For validation and testing, we simultaneously exposed all patches of a slide to the model to avoid sampling biases and statistical instabilities. Due to the computational complexity of TransMIL, we excluded slides with more than 24,000 patches ($\approx 6\%$ of all slides). We did this for all methods to ensure fair comparisons. AttnMIL models were trained for up to 1,000 epochs with batch size 32, and the TransMIL models for up to 200 epochs with batch size 5. We selected the checkpoint with the highest validation AUC. In the HNSC HPV dataset, we used one fold as validation and test fold and two folds as training folds and repeated this procedure for all possible assignments of folds. We applied a grid search over learning rates and dropout schemes and selected the hyper-parameter settings with the highest mean validation AUCs over 5 repetitions. For AttnMIL, we found that the best configuration was always a learning rate of 0.002 and no dropout. For TransMIL, we ended up with a learning rate of 0.0002 and high dropout (0.2 after the feature extractor, 0.5 after the self-attention blocks and before the final classification layer) for CAMELYON16 and NSCLC, and a learning rate of 0.002 without dropout for HNSC HPV and LUAD TP53. The training was done on an A100 80GB GPU.
\subsection{Faithfulness evaluation: Patch flipping} \label{app:perturbation_details}

Given a slide $X=\{\mathbf{x}_k\}_{k=1}^{K}$ and a heatmaping function $\mathcal{H}$ producing the explanation scores of the instances in $X$, i.e., $\mathcal{H}(\mathbf{x}_k)=\hat{\epsilon}_{k}$, we binned the patches of slide $X$ into 100 ordered groups $(E_1, \cdots, E_{100})$, where $E_i$ is the set of all patches whose attribution scores are between the $(100-i)$-th and $(100-i+1)$-th percentiles of the explanation scores of the instances of $X$, for example, $E_1$ is the set of the most relevant 1\% patches of $X$ and $E_{100}$ is the least relevant 1\% of the patches.

\textbf{Patch dropping}. Following the region perturbation strategy introduced in \cite{samek2016evaluating}, we progressively excluded the most relevant regions from slide $X$, i.e. in the $n$-th iteration, the most relevant $n$\% of patches were excluded. Formally, the perturbation procedure can be formulated as the following:

\begin{equation}\label{eq:dropping_morf}
\begin{gathered}
X_{\text{morf}}^{(0)} = X \\
X_{\text{morf}}^{(n)} = \mathcal{P}(X, n) \\
\mathcal{P}(X, n) = \bigcup_{i=n+1}^{100}E_i
\end{gathered}
\end{equation}
where $\mathcal{P}(X, n)$ is a perturbation function that excludes the most relevant $n$\% of patches from slide $X$. Note that at step $n=100$ all the patches are excluded and therefore, $X_{\text{morf}}^{(100)}=\emptyset$, where $\emptyset$ is an empty set, for which we pass an array of zeros to the model.

\textbf{Comparing heatmaps}. We define the quantity of interest for the comparison of different heatmaps as the area under the perturbation curve (AUPC) as the following:
\begin{equation} \label{eq:aupc}
    \text{AUPC}(X, \mathcal{H}) = \frac{1}{100} \sum_{n=0}^{101} f(X_{{\text{morf}}}^{(n)})
\end{equation}
where $f$ is the model's function.

Heatmap $\mathcal{H}_1$ is more faithful than $\mathcal{H}_2$ if $ \text{AUPC}(X, \mathcal{H}_1) <  \text{AUPC}(X, \mathcal{H}_2)$. That is, the lower AUPC, the more faithful the explanation method.

We ran all the AttnMIL experiments on an A100 40GB GPU and the TransMIL experiments on a single CPU.
\subsection{Faithfulness evaluation: Additional results} \label{app:perturbation_results}

\begin{figure}[h]
    \centering
    \includegraphics[width=0.95\textwidth]{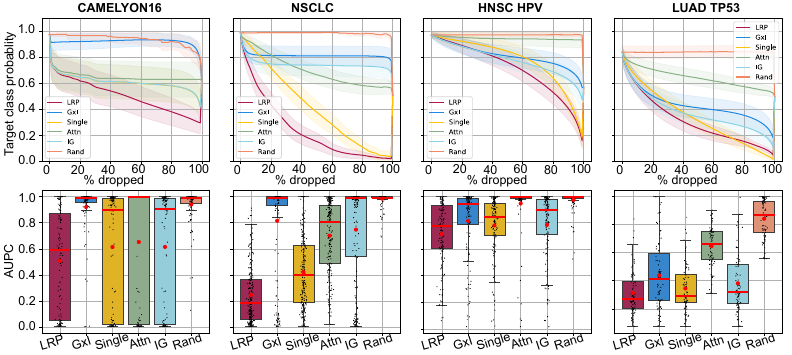}
    \caption{Patch dropping results for AttnMIL. The first row depicts the perturbation curves, where the solid lines are the average perturbation curve and the shaded area is the standard error of the mean at each perturbation step. Each boxplot on the second row shows the distribution of AUPC values for all test set slides per explanation methods. In each boxplot, the red line marks the median and the red dot marks the mean. \emph{Lower perturbation curves and AUPCs represent higher faithfulness.}}
    \label{fig:attnmil_dropping_results}
\end{figure}

We present the patch dropping results for AttnMIL in Figure \ref{fig:attnmil_dropping_results}.

\subsection{Extracting insights from \method heatmaps: Additional results} \label{app:additional_heatmaps}

\begin{figure}[h!]
    \centering
    \includegraphics[width=\textwidth]{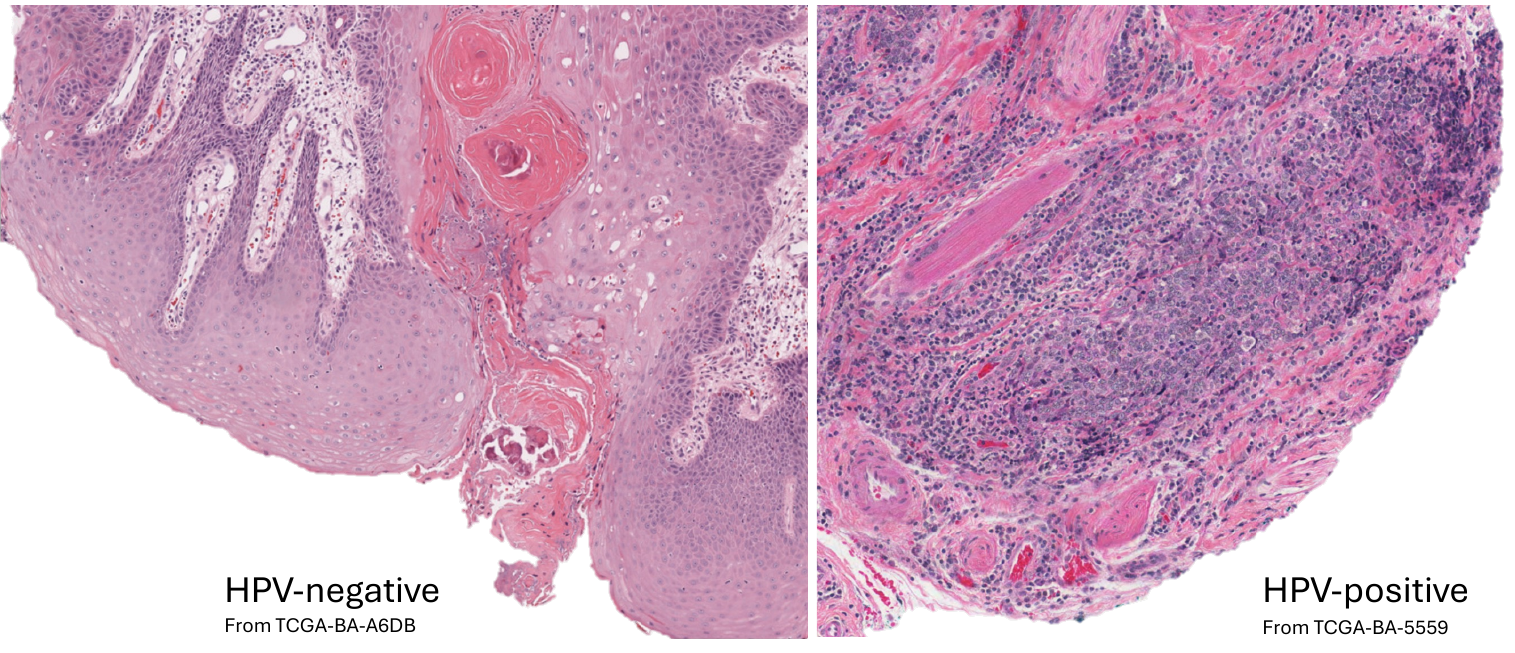}
    \caption{Exemplary histological features of HPV-negative and -positive HNSC.}
    \label{fig:hnscc_hpv_features}
\end{figure}

Figure \ref{fig:hnscc_hpv_features} summarizes and compares the appearance (pathologists call this morphology or histological features) of HPV-negative and HPV-positive HNSCC. HPV-positive HNSCC generally do not produce keratin (non-keratinizing morphology) and unlike HPV-negative do not origin from the surface epithelium but crypt epithelium of palatine and lingual tonsils. HPV-positive tumor nests are often embedded in lymphocyte rich parts of the tissue (lymphoid stroma) and the tumor nuclei show a darker, denser staining (i.e. hyperchromatic nuclei).

\begin{figure}[h!]
    \centering
    \includegraphics[width=0.8\textwidth]{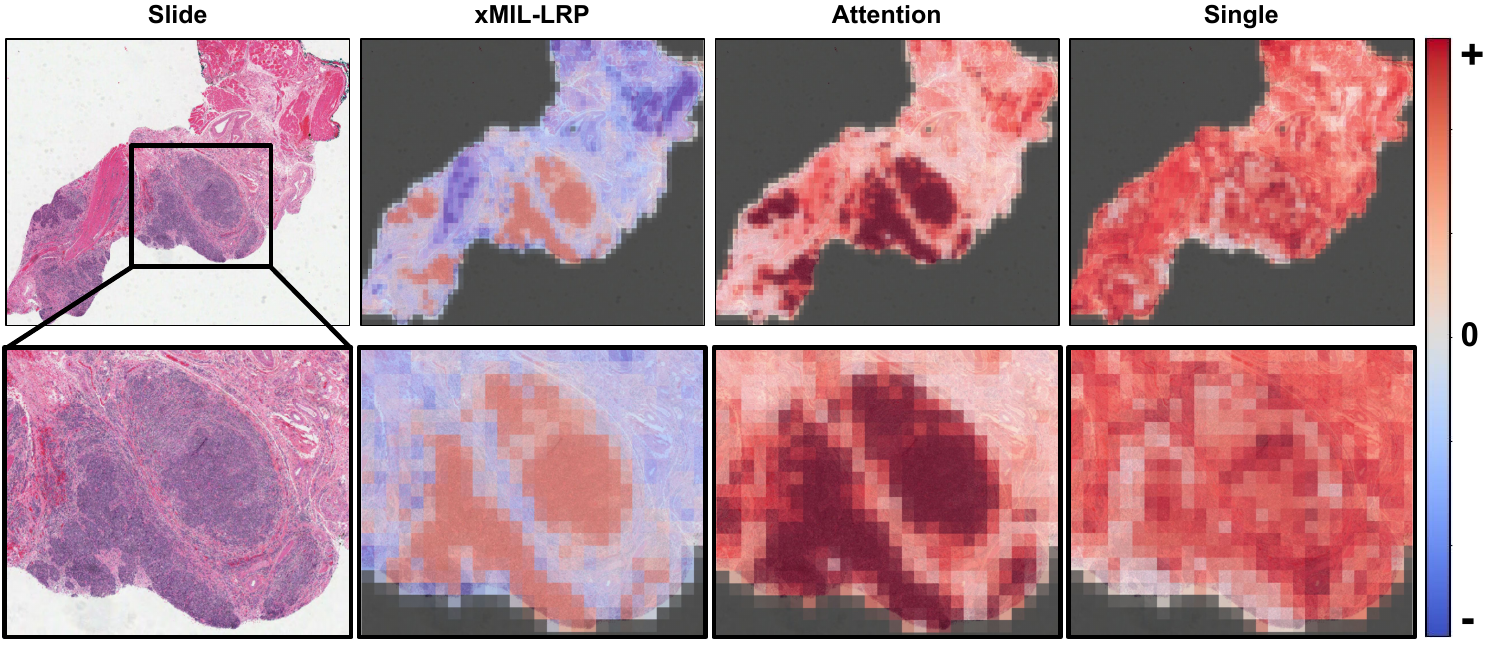}
    \caption{Heatmaps from different explanation methods for a TransMIL model predicting HPV-status. The model correctly predicted the slide HPV-positive (prediction score: 0.9215). For \methodnospace, red indicates evidence for and blue against the HPV-positive class.}
    \label{fig:hnscc_heatmap_3}
\end{figure}

\begin{figure}[h!]
    \centering
    \includegraphics[width=0.8\textwidth]{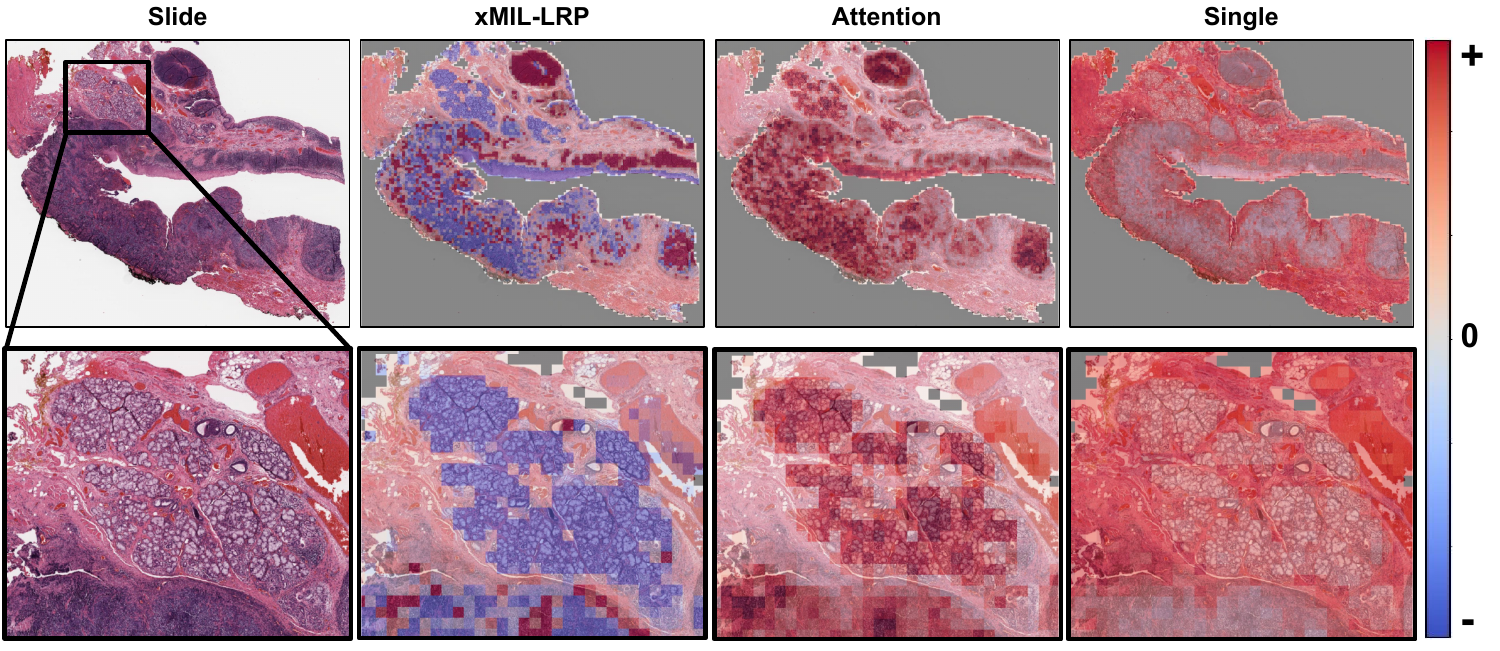}
    \caption{Heatmaps from different explanation methods for a TransMIL model predicting HPV-status. The model correctly predicted the slide HPV-positive (prediction score: 0.9048). For \methodnospace, red indicates evidence for and blue against the HPV-positive class.}
    \label{fig:hnscc_heatmap_1}
\end{figure}

\begin{figure}[h!]
    \centering
    \includegraphics[width=0.8\textwidth]{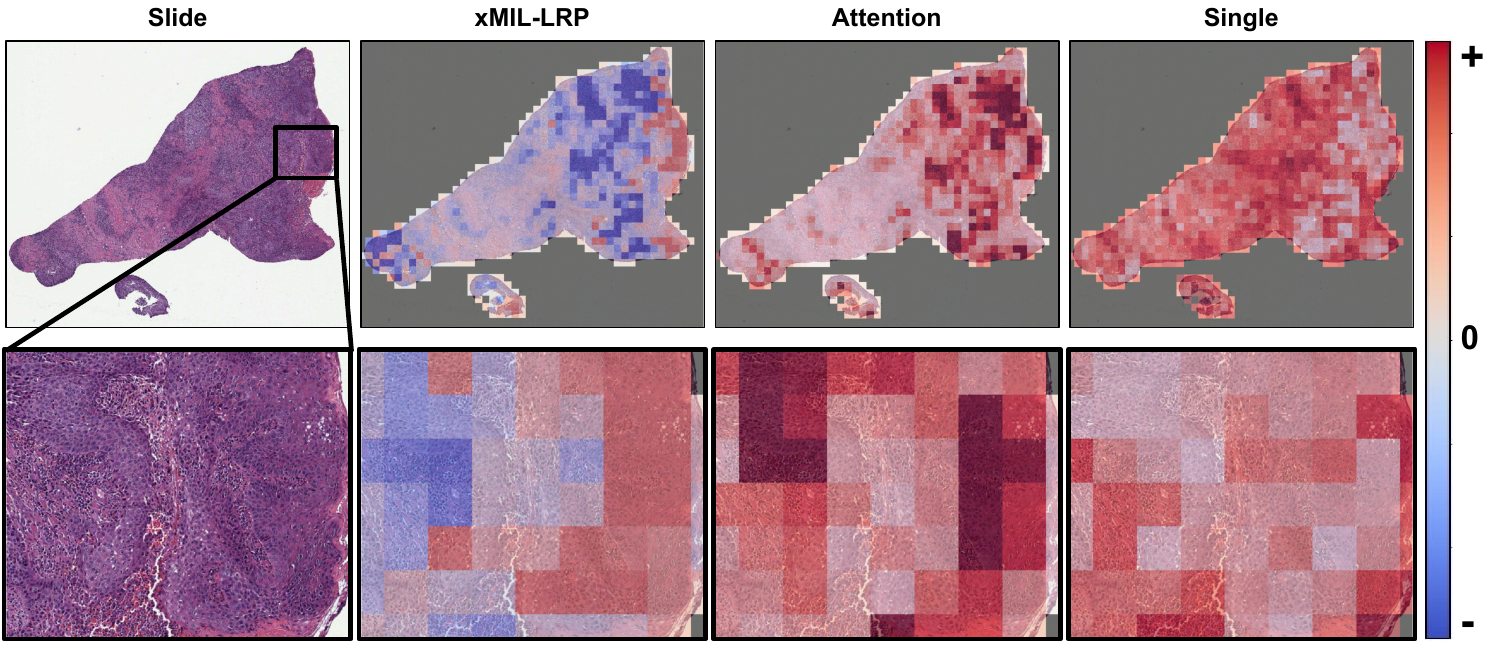}
    \caption{Heatmaps from different explanation methods for a TransMIL model predicting HPV-status. The slide is HPV-negative, but the model predicted HPV-positive (prediction score: 0.9997). For \methodnospace, red indicates evidence for and blue against the HPV-negative class.}
    \label{fig:hnscc_heatmap_4}
\end{figure}

We display further exemplary heatmaps of TransMIL model predictions in the HNSC HPV dataset in Figures \ref{fig:hnscc_heatmap_3}, \ref{fig:hnscc_heatmap_1}, and \ref{fig:hnscc_heatmap_4}.

\end{document}